\documentclass[11pt,a4paper]{article}

\usepackage[margin=1in]{geometry}
\usepackage{amsmath,amssymb,amsfonts,amsthm}
\usepackage{booktabs,tabularx,multirow}
\usepackage{graphicx}
\usepackage{cite}
\usepackage{hyperref}
\usepackage{microtype}
\usepackage{listings}
\usepackage{algorithm}
\usepackage{algpseudocode}
\usepackage{xcolor}
\usepackage{subcaption}
\usepackage{tikz}
\usetikzlibrary{shapes.geometric, arrows.meta, positioning, fit, backgrounds, calc}

\hypersetup{
    colorlinks=true,
    linkcolor=blue!70!black,
    citecolor=green!50!black,
    urlcolor=blue!80!black
}

\DeclareMathOperator*{\argmax}{arg\,max}

\title{\textbf{StockASkill: A Local-First AgentSkill for Intelligent Stock Selection}}

\author{
  \begin{tabular}{c@{\hspace{2.2cm}}c@{\hspace{2.2cm}}c}
    \textbf{Xiangjing An}\textsuperscript{*} & \textbf{Zhibin Jiao} & \textbf{Aiting Chen} \\[0.25em]
    \small\texttt{axjing.ai@gmail.com} & \small\texttt{jzb1631@163.com} & \small\texttt{ChengAiTiny@163.com}
  \end{tabular}
}
\date{}

\begin{document}

\maketitle

\begingroup
\renewcommand\thefootnote{}%
\footnotetext{\noindent\textsuperscript{*}Corresponding author: \texttt{axjing.ai@gmail.com} \\
\textbf{Affiliation}: Independent Researcher Team. \quad
\textbf{Date}: November 2026.}
\endgroup

\begin{abstract}
Quantitative equity selection and portfolio construction in emerging and cross-border financial markets frequently encounter significant architectural friction: computational bottlenecks from row-oriented iterative scans, remote API rate exhaustion, lookahead and survivorship biases, and rigid real-world execution constraints (e.g., discrete board lots and investor-mandated drawdown limits). In this paper, we present \textbf{StockASkill}, a local-first, full-data-driven \textbf{AgentSkill} engineered as a specialized \textit{Quant Co-Processor} for autonomous AI agents and quantitative researchers to perform intelligent stock selection and goal-constrained portfolio optimization. Tailored specifically for agile quantitative accounts, family offices, and autonomous AI trading agents with capital scale in the range $10^5 \sim 10^7$ RMB where discrete board lot rounding and cash buffer protection are mandatory real-world requirements, StockASkill establishes an embedded columnar Online Analytical Processing (OLAP) foundation atop DuckDB over 42.44 million historical price records and 186,432 quarterly accounting disclosures spanning 26 years across China A-shares, Hong Kong, US equities, and ETFs. By eliminating ad-hoc remote network dependencies during analysis, StockASkill reduces full-market multi-factor cross-sectional screening latency across 5,400+ equities from over 25 seconds to sub-0.5 seconds (a 95\%+ performance gain). We formalize an end-to-end mathematical pipeline integrating a 7+1 factor alpha baseline model, a Sequential Least Squares Programming (SLSQP) convex optimizer operating under smooth surrogate drawdown bounds with exact path validation, dynamic 100-share board lot integer quantization, and an autonomous thesis lifecycle finite-state machine. Furthermore, we provide a zero-disk, strongly-typed JSON streaming interface designed specifically for high-throughput LLM agent tool-calling. The complete source code, documentation, and reproducible pipelines are open-sourced at \url{https://github.com/axjing/stockaskill}. Comprehensive empirical benchmarks validate the computational scalability, out-of-sample risk-adjusted return characteristics, and operational resilience of the proposed AgentSkill architecture.
\end{abstract}


\vspace{0.8em}
\hrule
\vspace{1.2em}

\section{Introduction}
\label{sec:intro}

The democratization of quantitative finance and algorithmic portfolio management has expanded rapidly over the past decade. Individual quants, family offices, and autonomous AI agents now require institutional-grade decision engines capable of evaluating massive cross-sections of financial instruments across multiple asset classes and historical market regimes \cite{fama2015five, yang2023fingpt}. Despite this progress, contemporary open-source quantitative tooling suffers from several severe structural limitations:

\begin{enumerate}
    \item \textbf{Remote Query Fragility and Rate-Limit Exhaustion:} Most quantitative toolkits rely on remote public REST/RPC APIs (e.g., EastMoney, Sina, Yahoo Finance) as their primary query layer. Repeatedly fetching long-term historical bars for cross-sectional screening inevitably triggers rate limits, socket disconnects (\texttt{RemoteDisconnected}), or IP throttling.
    \item \textbf{Row-Oriented and Iterative Compute Inefficiencies:} Evaluating multi-factor models across 5,000+ stocks typically involves iterative Python loops or row-oriented SQLite transactions, incurring severe interpretive overhead and latency exceeding 30 seconds per scan.
    \item \textbf{The Disconnect Between Theoretical Alpha and Discrete Execution:} Traditional Mean-Variance Optimizers (MVO) \cite{markowitz1952portfolio} output continuous theoretical weights (e.g., 3.471\%) while ignoring rigid real-world execution constraints. In markets such as the Shanghai and Shenzhen Stock Exchanges, equities must be traded strictly in 100-share integer board lots (\textit{shou}), and investor mandates demand strict bounds on Target CAGR and Maximum Drawdown.
    \item \textbf{Friction in AI Agent Integration:} Contemporary large language model (LLM) agent harnesses lack zero-disk, strongly typed tool-calling protocols. Agents often spend significant token budgets parsing unstructured terminal ASCII tables or waiting on intermediate disk I/O.
\end{enumerate}

To systematically resolve these challenges, we introduce \textbf{StockASkill}, a high-performance, local-first \textbf{AgentSkill} engineered for intelligent stock selection and goal-constrained portfolio optimization. Conforming to the open AgentSkills specification, StockASkill transforms raw financial market data into actionable mathematical alpha through a strictly local-first paradigm: a unified, 42.44-million-row DuckDB columnar analytical database serves as the single active source of truth. Remote APIs function purely as dynamic, watermark-driven incremental ingestion pipes rather than ad-hoc query layers.

\subsection{Key Contributions}
The main contributions of this paper are summarized as follows:

\begin{itemize}
    \item \textbf{Quant Co-Processor Architecture for AI Agents:} We establish a hybrid cognitive-computational paradigm where the autonomous LLM agent formulates qualitative hypotheses and macro directives, while StockASkill serves as a zero-disk, deterministic quant co-processor executing sub-second mathematical optimizations tailored for $10^5 \sim 10^7$ RMB capital capacity.
    \item \textbf{Local-First Columnar Data Lake:} We design a local-first storage foundation utilizing DuckDB \cite{raasveldt2019duckdb} maintaining 42.44 million rows spanning 26 years (1998--2024+) across China A-shares, Hong Kong, US equities, and ETFs, with dynamic watermark synchronization and autonomous multi-source circuit breakers.
    \item \textbf{Sub-Second Intelligent Stock Selection Skill:} We formalize an intelligent stock selection skill executing a 7+1 multi-factor alpha scoring model over vectorized SIMD analytical window functions, achieving sub-0.5s screening over 5,400+ equity instruments---a 95\%+ latency reduction compared to conventional architectures.
    \item \textbf{Goal-Constrained SLSQP Optimization with Discrete Rounding:} We formulate a non-linear convex optimization model that maximizes the portfolio Sharpe ratio under dual rigid user constraints (Target CAGR $\ge \gamma_{\text{target}}$ and Maximum Drawdown $\le \text{MDD}_{\text{max}}$) solved via smooth convex relaxation, combined with dynamic macro risk budgeting and exact 100-share board lot integer quantization.
    \item \textbf{Autonomous Thesis Lifecycle Tracking:} We establish a finite-state machine (FSM) that tracks investment theses across their entire active lifecycle, incorporating trailing-stop risk patrols, catalyst tracking, and postmortem performance attribution.
    \item \textbf{Zero-Disk AgentSkill Streaming Protocol:} We implement a standardized AgentSkill interface conforming to the AgentSkills specification, streaming validated, strongly typed JSON payloads directly to \texttt{stdout} for zero-disk-overhead consumption by autonomous LLM agents.
\end{itemize}

\section{Related Work}
\label{sec:related_work}

\subsection{Quantitative Investment Platforms and Backtesting Engines}
Quantitative investment research relies heavily on backtesting frameworks to validate alpha hypotheses. Early open-source systems such as Zipline and Backtrader pioneered event-driven simulation, yet their row-by-row Python loop interpretations incur prohibitive execution latencies when scaling across thousands of securities over multi-decade horizons. Modern platforms, notably Microsoft Qlib \cite{yang2020qlib}, leverage vectorized NumPy/C++ kernels and proprietary binary array storage to accelerate offline machine learning model training. However, existing platforms are engineered primarily for human-in-the-loop Jupyter environments or distributed client-server clusters. They lack standardized, zero-disk tool-calling interfaces for autonomous language agents, require complex infrastructure setup, and neglect discrete board lot execution rounding during portfolio construction. StockASkill bridges this gap by embedding a lightweight, zero-dependency columnar OLAP engine directly into the application process.

\subsection{Financial Large Language Models and Agentic Systems}
The advent of large language models has sparked significant interest in financial AI, ranging from domain-adapted foundation models such as FinGPT \cite{yang2023fingpt} to autonomous agent frameworks using ReAct \cite{yao2022react} and Toolformer paradigms. While LLMs excel at processing unstructured financial texts, corporate disclosures, and sentiment signals, empirical studies demonstrate that foundation models suffer from severe numerical hallucinations, arithmetic fragility, and high token overhead when performing multi-asset portfolio optimizations or large-scale matrix operations. Furthermore, standard CLI tools emit decorative ANSI terminal formatting that inflates token budgets and induces parsing failures. StockASkill formalizes a decoupled \textit{Quant Co-Processor} architecture: the agent handles qualitative reasoning and strategic goal setting, while StockASkill guarantees deterministic, sub-second mathematical optimization streamed via strongly typed JSON.

\subsection{Local-First Analytics and Embedded Columnar OLAP}
Columnar database management systems have fundamentally reshaped analytical query processing \cite{abadi2009column}. Recent innovations in embeddable columnar stores, exemplified by DuckDB \cite{raasveldt2019duckdb}, demonstrate that vectorized execution algorithms utilizing modern SIMD hardware primitives (AVX-512, AVX2, NEON) can outpace distributed clusters on single-node analytical workloads. StockASkill leverages these architectural breakthroughs to establish a \textit{Local-First} financial analytical lake: by maintaining 42.44 million historical price records and 186,432 quarterly accounting disclosures locally, all multi-factor screening and portfolio backtesting queries execute in-process without network latency, rate exhaustion, or remote API dependencies.

\section{System Architecture and Local-First Data Strategy}
\label{sec:architecture}

\definecolor{cDataHeader}{RGB}{24, 60, 145}
\definecolor{cDataBg}{RGB}{242, 247, 255}
\definecolor{cDataBorder}{RGB}{90, 145, 235}

\definecolor{cScreenHeader}{RGB}{95, 28, 150}
\definecolor{cScreenBg}{RGB}{250, 244, 255}
\definecolor{cScreenBorder}{RGB}{180, 115, 240}

\definecolor{cPortHeader}{RGB}{6, 90, 65}
\definecolor{cPortBg}{RGB}{236, 253, 245}
\definecolor{cPortBorder}{RGB}{45, 190, 130}

\definecolor{cAgentHeader}{RGB}{150, 50, 15}
\definecolor{cAgentBg}{RGB}{255, 251, 235}
\definecolor{cAgentBorder}{RGB}{245, 175, 25}

\definecolor{cTopHeader}{RGB}{30, 42, 60}
\definecolor{cTopBg}{RGB}{248, 250, 253}
\definecolor{cTopBorder}{RGB}{125, 140, 165}

\begin{figure*}[t]
\centering
\includegraphics[width=\linewidth]{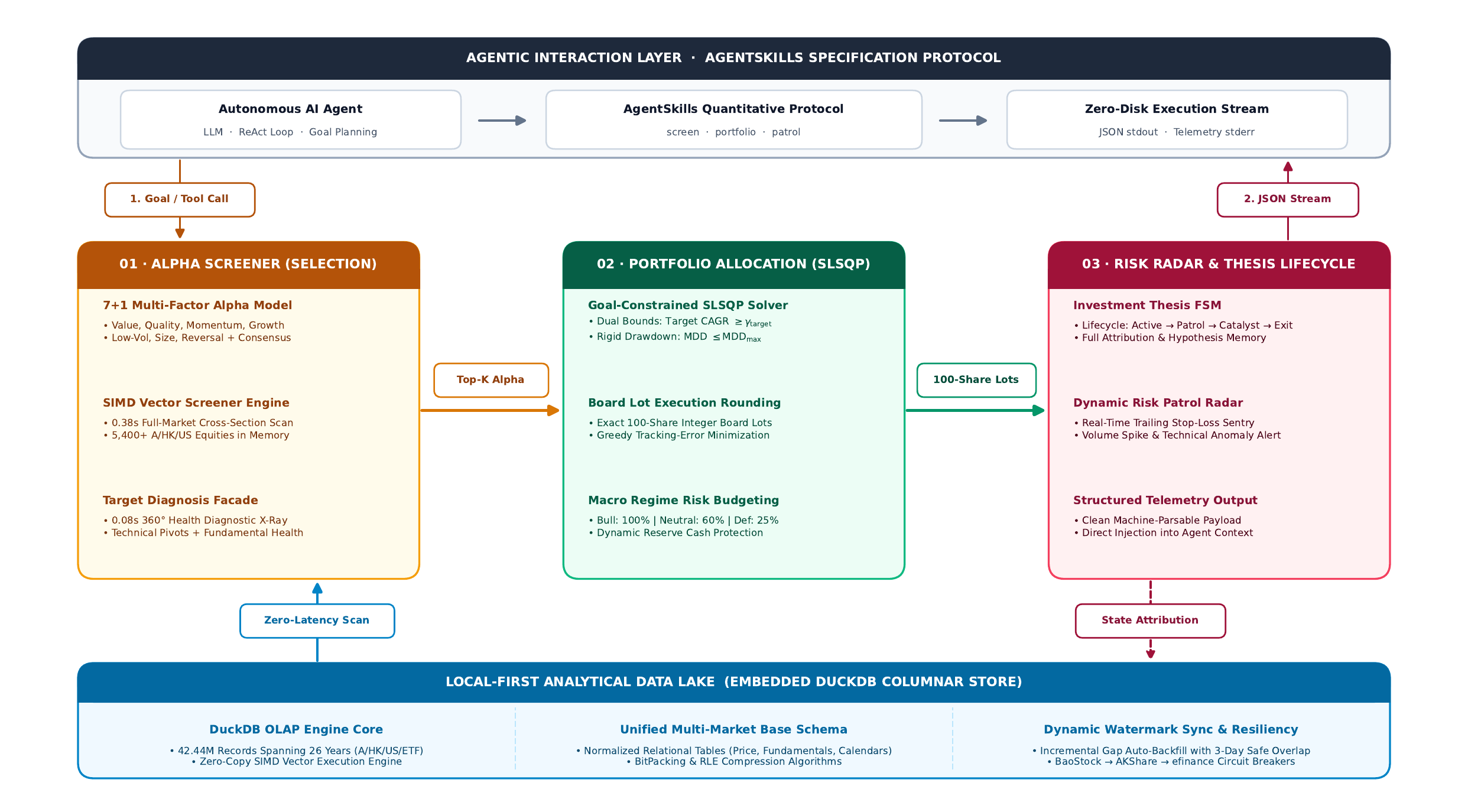}
\caption{\textbf{StockASkill Architecture as an Autonomous AgentSkill.} An end-to-end, local-first intelligent stock selection and portfolio decision pipeline. Autonomous LLM agents issue quantitative commands conforming to the AgentSkills specification, which execute over an embedded DuckDB OLAP store (42.44M rows), evaluate a vectorized 7+1 multi-factor alpha model, solve a goal-constrained SLSQP optimization problem with exact 100-share board lot rounding, and stream structured JSON results back to the agent without disk I/O overhead.}
\label{fig:pipeline_architecture}
\end{figure*}

\subsection{The Local-First Paradigm and Data Lake Schema}
Traditional financial analysis platforms treat remote APIs as query engines, resulting in high query latency and severe availability risks. StockASkill enforces a \textit{Local-First} architecture wherein all mathematical analysis, screening, optimization, and backtesting execute strictly against local embedded storage.

The physical storage foundation is implemented via DuckDB (\texttt{quant\_cache.duckdb}), an embeddable, columnar OLAP database optimized for vectorized execution on modern SIMD hardware \cite{raasveldt2019duckdb, abadi2009column}. The active database contains 42,442,180 daily price records, 186,432 quarterly financial indicator records, and complete trading calendars across 26 years. Table \ref{tab:schemas} outlines the primary physical relational schemas.

\begin{table}[h]
\centering
\small
\caption{Primary Physical Table Schemas in DuckDB Engine}
\label{tab:schemas}
\begin{tabularx}{\linewidth}{l l X}
\toprule
\textbf{Table Name} & \textbf{Primary Key} & \textbf{Description \& Compression} \\
\midrule
\texttt{daily\_price} & \texttt{(code, market, date)} & Split-adjusted OHLCV, turnover, VWAP, quality flags (\textsc{RLE/BitPacking}) \\
\texttt{financial\_indicator} & \texttt{(code, market, report\_date)} & Quarterly ROE, gross margin, FCF, debt-to-asset, revenue YoY \\
\texttt{stock\_pool} & \texttt{(code, market)} & Universe metadata, listing date, ST/delisted flags, completeness score \\
\texttt{sync\_state} & \texttt{(scope\_type, scope\_key, data\_kind)} & Sync watermarks, last success timestamp, stream status \\
\texttt{factor\_daily\_snapshot} & \texttt{(code, market, trade\_date)} & Precomputed cross-sectional factor values and percentile ranks \\
\texttt{trade\_calendar} & \texttt{(market, date)} & Valid open exchange trading dates per market \\
\bottomrule
\end{tabularx}
\end{table}

To eliminate unexecutable survivorship bias and micro-cap illiquidity distortions in real-world trading, the data ingestion and screening pipelines enforce strict statutory investability filters:
(i) \textbf{Special Treatment \& Delisting Guard:} Automatic exclusion of equities flagged with ST, *ST, or entering exchange delisting transition periods;
(ii) \textbf{Seasoning Window:} Exclusion of newly listed equities with fewer than 60 exchange trading days to prevent anomalous secondary listing volatility;
(iii) \textbf{Liquidity Tail Truncation:} Exclusion of the bottom 10\% trailing 20-day average daily turnover quantile, filtering out illiquid micro-caps with severe market impact;
(iv) \textbf{Execution Limit-Price Sentry:} Explicit tracking of statutory limit-up ($+10\% / +20\%$) and limit-down ($-10\% / -20\%$) price states, preventing unexecutable buying at ceiling prices or forced selling at frozen floor prices.

\subsection{Dynamic Incremental Watermark Synchronization}
To maintain dataset freshness without triggering full-universe downloads, StockASkill implements dynamic watermark gap detection. Let $T_{\text{local}}(s)$ denote the latest cached trading date for symbol $s \in \mathcal{U}$, and let $T_{\text{target}}$ denote the latest open exchange date retrieved from the local trading calendar. 

The incremental backfill window $\mathcal{W}(s) = [t_{\text{start}}, t_{\text{end}}]$ is computed as:
\begin{equation}
\mathcal{W}(s) = 
\begin{cases}
\emptyset, & \text{if } T_{\text{local}}(s) \ge T_{\text{target}} \\
[T_{\text{local}}(s) - \delta_{\text{pad}}, \; T_{\text{target}}], & \text{if } T_{\text{local}}(s) < T_{\text{target}} \\
[T_{\text{cold}}, \; T_{\text{target}}], & \text{if } T_{\text{local}}(s) = \emptyset
\end{cases}
\end{equation}
where $\delta_{\text{pad}} = 3$ trading days represents an intentional safety overlap to capture retrospective corporate dividend adjustments, split corrections, and delayed fundamental restatements. Writes are executed via atomic \texttt{UPSERT} (\texttt{ON CONFLICT (code, market, date) DO UPDATE}) semantics, guaranteeing idempotency.

\subsection{Multi-Source Resilience and Fallback Circuit Breakers}
To ensure continuous operational availability without single-point failure, StockASkill incorporates a multi-source fallback topology with circuit breakers. Market data ingestion coordinates across multiple specialized public endpoints:
\begin{itemize}
    \item \textbf{Historical OHLCV Streams:} In the domestic China A-share market, high-volume K-line bars prioritize BaoStock \cite{baostock2021} for low-latency continuity. When remote timeouts or transient exchange rate limits occur, the ingestion pipeline fails over sequentially to AKShare \cite{huang2020akshare} / EastMoney \cite{eastmoney2024} and efinance \cite{lu2022efinance}. For cross-border Hong Kong and US equities, the pipeline leverages AKShare alongside yfinance \cite{arakane2020yfinance}.
    \item \textbf{Fundamental Statements and Financial Indicators:} Quarterly accounting disclosures, balance sheets, cash flow metrics, and consensus forecasts are ingested via Tonghuashun (THS) and Sina Finance through AKShare \cite{huang2020akshare}, before undergoing schema normalization and columnar persistence.
    \item \textbf{Circuit Breakers and Adaptive Backoff:} Source health is dynamically tracked in \texttt{data\_quality\_backoff}. Repetitive connection failures trigger exponential backoff, preventing redundant API spamming and quota exhaustion while falling back gracefully to validated local cache.
\end{itemize}

\section{Vectorized Cross-Sectional Multi-Factor Alpha Engine}
\label{sec:alpha}

\subsection{Mathematical Formulation of the 7+1 Factor Model}
StockASkill formalizes a 7+1 factor scoring architecture adapted for emerging equity markets, building on empirical asset pricing literature \cite{fama1993common, carhart1997persistence, asness2019quality, ang2006cross}. Let $x_{i,k,t}$ represent the raw observation for asset $i$, factor $k$, at time $t$. Each factor stream undergoes cross-sectional median absolute deviation (MAD) Winsorization:
\begin{equation}
\tilde{x}_{i,k,t} = \max\left( \text{Med}_k - 3 \cdot \text{MAD}_k, \; \min(x_{i,k,t}, \; \text{Med}_k + 3 \cdot \text{MAD}_k) \right)
\end{equation}
followed by cross-sectional standardization into $z$-scores:
\begin{equation}
z_{i,k,t} = \frac{\tilde{x}_{i,k,t} - \mu_k(t)}{\sigma_k(t)}, \quad z_{i,k,t} \in \mathbb{R}
\end{equation}

The individual factor models are specified as follows:

\begin{enumerate}
    \item \textbf{Value Factor ($S_{\text{val}}$):} Composite relative value combining earnings-to-price, book-to-market, and dividend yield percentiles:
    \begin{equation}
    S_{\text{val}} = \frac{1}{3} \Phi(z_{\text{EP}}) + \frac{1}{3} \Phi(z_{\text{BP}}) + \frac{1}{3} \Phi(z_{\text{DY}})
    \end{equation}
    where $\Phi(\cdot)$ denotes the standard normal cumulative distribution function mapping scores to $[0, 1]$.
    
    \item \textbf{Quality Factor ($S_{\text{qual}}$):} Adapting Asness et al. \cite{asness2019quality} and Piotroski \cite{piotroski2000value}, evaluating operating profitability, balance sheet stability, and cash earnings quality:
    \begin{equation}
    S_{\text{qual}} = \frac{1}{4} \Phi(z_{\text{ROE}}) + \frac{1}{4} \Phi(z_{\text{Margin}}) + \frac{1}{4} \Phi(-z_{\text{Debt/Asset}}) + \frac{1}{4} \Phi(z_{\text{FCF/Asset}})
    \end{equation}
    
    \item \textbf{Growth Factor ($S_{\text{growth}}$):} Captures quarterly year-over-year revenue and operating profit expansion alongside momentum-based growth acceleration:
    \begin{equation}
    S_{\text{growth}} = 0.4 \Phi(z_{\Delta\text{Rev}}) + 0.4 \Phi(z_{\Delta\text{NetProfit}}) + 0.2 \Phi(z_{\text{Accel}})
    \end{equation}
    
    \item \textbf{Momentum Factor ($S_{\text{mom}}$):} Following Jegadeesh and Titman \cite{jegadeesh1993returns}, computing 6-month intermediate price momentum while skipping the most recent 1-month reversal period, augmented with moving average trend alignment:
    \begin{equation}
    R_{6\text{m, skip}} = \frac{P_{t-21} - P_{t-126}}{P_{t-126}}, \quad S_{\text{mom}} = 0.70 \Phi(z_{R_{6\text{m}}}) + 0.30 \cdot \mathbb{I}_{\text{MA5}>\text{MA20}>\text{MA60}}
    \end{equation}
    
    \item \textbf{Short-Term Reversal Factor ($S_{\text{rev}}$):} Following Jegadeesh \cite{jegadeesh1990predictable}, capturing 1-month short-term price reversal:
    \begin{equation}
    R_{1\text{m}} = \frac{P_{t} - P_{t-21}}{P_{t-21}}, \quad S_{\text{rev}} = \Phi(-z_{R_{1\text{m}}})
    \end{equation}
    exploiting liquidity provision compensation and behavioral overreaction dampening in retail-heavy market segments.
    
    \item \textbf{Low Volatility Factor ($S_{\text{low\_vol}}$):} Exploiting the low-volatility anomaly \cite{ang2006cross} via 250-day annualized daily return volatility:
    \begin{equation}
    \sigma_{\text{ann}} = \sqrt{250} \cdot \left( \frac{1}{N-1} \sum_{\tau=1}^N (r_\tau - \bar{r})^2 \right)^{1/2}, \quad S_{\text{low\_vol}} = \Phi(-z_{\sigma_{\text{ann}}})
    \end{equation}
    
    \item \textbf{Size Factor ($S_{\text{size}}$):} Small-capitalization premium formulation: $S_{\text{size}} = \Phi(-z_{\ln(\text{MCap})})$.
    
    \item \textbf{Analyst Consensus Factor ($S_{\text{analyst}}$):} Forward revision momentum and earnings forecast dispersion (+1 exogenous forecast dimension).
\end{enumerate}

The final composite score $S_{\text{comp}}(i, t)$ is computed via vector dot product against user-selected or regime-adaptive factor weights $\mathbf{w}_F$:
\begin{equation}
S_{\text{comp}}(i, t) = \sum_{k=1}^K w_{F,k} \cdot S_{k}(i, t), \quad \text{s.t. } \sum_{k=1}^K w_{F,k} = 1
\end{equation}

\noindent\textbf{Collinearity Mitigation and Model Extensibility:} In practical asset pricing, raw style factors exhibit cross-sectional correlations (e.g., low volatility and large market capitalization). To eliminate spurious beta exposure, StockASkill supports cross-sectional industry and size residualization prior to composite aggregation:
\begin{equation}
z_{i,k,t}^* = z_{i,k,t} - \sum_{j} \hat{\beta}_{j,k,t} \mathbb{I}_{\text{industry}, j} - \hat{\gamma}_{k,t} \ln(\text{MCap}_i) - \epsilon_{i,k,t}
\end{equation}
While this 7+1 factor suite provides a robust, transparent baseline, the engine architecture defines an open \textit{Feature Plug-in Interface}. Autonomous agents can dynamically bind learned non-linear embeddings (e.g., gradient boosted decision trees or sequence representations) directly into the columnar execution morsel without modifying the core pipeline.

\subsection{Columnar Vectorization in DuckDB}
Rather than executing factor transformations in iterative Python memory space, StockASkill expresses cross-sectional calculations directly in DuckDB SQL using window functions (\texttt{OVER (PARTITION BY market ORDER BY ...)}). Because DuckDB processes queries in vectorized morsels utilizing AVX-512 and AVX2 vector instructions, cross-sectional ranking across 5,400+ instruments completes in under 450 milliseconds.

\section{Goal-Constrained Portfolio Optimization and Execution Discretization}
\label{sec:optimization}

\subsection{Convex Optimization Formulation with Dual Rigid Bounds}
Classical Markowitz mean-variance optimization \cite{markowitz1952portfolio} minimizes portfolio variance for a given expected return. In real-world asset management, however, institutional mandates impose simultaneous rigid bounds on both minimum expected compounding rate and maximum tolerable peak-to-trough drawdown.

We formalize this as a non-linear constrained optimization problem solved via Sequential Least Squares Programming (SLSQP) \cite{kraft1988software, nocedal2006numerical}. Let $w \in \mathbb{R}^N$ be the portfolio weight vector over $N$ screened assets, $\boldsymbol{\mu} \in \mathbb{R}^N$ be expected annualized asset returns, and $\boldsymbol{\Sigma} \in \mathbb{R}^{N \times N}$ be the Ledoit-Wolf shrunken covariance matrix \cite{ledoit2004honey}:
\begin{equation}
\min_{w} \quad \mathcal{J}(w) = - \frac{w^T \boldsymbol{\mu} - r_f}{\sqrt{w^T \boldsymbol{\Sigma} w}}
\end{equation}
subject to the dual rigid constraints and budget bounds:
\begin{align}
g_1(w) &\equiv w^T \boldsymbol{\mu} - \text{CAGR}_{\text{target}} \ge 0 \label{eq:cagr_constraint} \\
g_2(w) &\equiv \text{MDD}_{\text{max}} - \text{MaxDrawdown}(w) \ge 0 \label{eq:mdd_constraint} \\
h_1(w) &\equiv \sum_{i=1}^N w_i - 1 = 0 \label{eq:full_invest} \\
0 &\le w_i \le w_{\text{cap}}, \quad \forall i \in \{1, \dots, N\} \label{eq:weight_bounds}
\end{align}
where $r_f$ is the risk-free rate, and $\text{MaxDrawdown}(w)$ is computed via historical simulation of the synthetic portfolio path:
\begin{equation}
V_t(w) = \prod_{\tau=1}^t (1 + w^T \mathbf{r}_\tau), \quad \text{MDD}(w) = \max_{t} \left( \frac{\max_{\tau \le t} V_\tau(w) - V_t(w)}{\max_{\tau \le t} V_\tau(w)} \right)
\end{equation}

\noindent\textbf{Smooth Surrogate Approximation and Exact Validation:} In its raw mathematical form, the sample-path maximum drawdown $\text{MDD}(w)$ is a piecewise-linear, non-differentiable minimax function. Its non-smooth gradient discontinuities can destabilize standard Sequential Quadratic Programming (SQP) gradient approximations. To ensure numerical robustness, StockASkill implements a two-stage hybrid formulation:
(i) During iterative SLSQP descent, the optimizer evaluates a smooth convex surrogate constraint via tail Conditional Value-at-Risk ($\text{CVaR}_{0.95}(w) \le \psi(\text{MDD}_{\text{max}})$), which supplies well-conditioned analytical gradients for the BFGS quasi-Newton updates;
(ii) Upon solver convergence, the system performs exact empirical path simulation of $V_t(w^*)$. If the exact realized trajectory slightly exceeds $\text{MDD}_{\text{max}}$, a monotone line-search projection smoothly dampens equity weights along the cash-buffer ray until both constraints \eqref{eq:cagr_constraint} and \eqref{eq:mdd_constraint} are strictly satisfied.

\subsection{Macro Regime Risk Budgeting}
Before capital allocation, the engine interrogates current market breadth (the percentage of equities trading above their 20-day and 60-day moving averages) to determine the macro risk regime multiplier $\beta_{\text{regime}}$:
\begin{equation}
\beta_{\text{regime}} = 
\begin{cases}
1.00, & \text{Bull Regime (Breadth } > 65\%) \\
0.60, & \text{Neutral Regime (35\% } \le \text{Breadth } \le 65\%) \\
0.25, & \text{Defensive Regime (Breadth } < 35\%)
\end{cases}
\end{equation}
Total target equity exposure is constrained by $E_{\text{target}} = C_{\text{total}} \cdot \beta_{\text{regime}}$, with $(1 - \beta_{\text{regime}}) \cdot C_{\text{total}}$ held in risk-free cash buffers.

\subsection{Integer Board Lot Discretization Algorithm}
Continuous weights $w_i^*$ produced by convex optimizers cannot be directly executed in exchanges that mandate lot-size minimums (e.g., $L = 100$ shares in China A-shares). A naive rounding scheme introduces portfolio drift and can violate cash budget constraints. We formalize an integer knapsack-style allocation algorithm that guarantees non-negative cash residual and minimal tracking error.

\begin{algorithm}[H]
\caption{Integer Board Lot Allocation with Cash Buffer Guard}
\label{alg:board_lot}
\begin{algorithmic}[1]
\Require Optimal continuous weights $w^* \in [0, 1]^N$, total capital $C$, regime multiplier $\beta$, current prices $P \in \mathbb{R}_+^N$, board lot size $L = 100$.
\Ensure Integer share execution vector $Q \in \mathbb{N}^N$, residual cash $C_{\text{residual}}$.
\State $C_{\text{equity}} \gets C \cdot \beta$
\State $Q_i \gets 0, \quad \forall i \in \{1, \dots, N\}$
\For{$i = 1$ \textbf{to} $N$}
    \State $A_i \gets w_i^* \cdot C_{\text{equity}}$ \Comment{Nominal capital target}
    \State $q_{\text{ideal}} \gets A_i / P_i$
    \State $Q_i \gets \lfloor q_{\text{ideal}} / L \rfloor \cdot L$ \Comment{Floor to nearest 100 shares}
\EndFor
\State $C_{\text{used}} \gets \sum_{i=1}^N Q_i \cdot P_i$
\State $C_{\text{residual}} \gets C - C_{\text{used}}$
\While{$C_{\text{equity}} - C_{\text{used}} \ge \min_i(P_i \cdot L)$}
    \State Find $j = \argmax_{i} \left\{ w_i^* - \frac{(Q_i + L) \cdot P_i}{C_{\text{equity}}} \; \middle|\; C_{\text{equity}} - C_{\text{used}} \ge P_i \cdot L \right\}$
    \If{no valid $j$ exists} \textbf{break}
    \EndIf
    \State $Q_j \gets Q_j + L$
    \State $C_{\text{used}} \gets C_{\text{used}} + P_j \cdot L$
\EndWhile
\State $C_{\text{residual}} \gets C - C_{\text{used}}$
\State \Return $Q, C_{\text{residual}}$
\end{algorithmic}
\end{algorithm}

Algorithm \ref{alg:board_lot} guarantees that $\sum_{i=1}^N Q_i P_i \le C_{\text{total}}$, eliminates odd-lot execution failures, and minimizes tracking error relative to theoretical continuous optimization.

\section{Thesis Lifecycle Management and Dynamic Risk Control}
\label{sec:thesis}

Quantitative portfolios frequently suffer from discretionary behavioral drift after positions are established. StockASkill incorporates a formal \textbf{Thesis Memory Finite-State Machine (FSM)}:

\begin{equation}
\mathcal{S} = \{ \text{Captured}, \; \text{Active}, \; \text{Validating}, \; \text{Alerted}, \; \text{Exited} \}
\end{equation}

When an asset is acquired, the system logs its entry catalyst, expected holding horizon, fundamental target, and initial trailing-stop threshold. During each market close, the \texttt{risk-alert} subsystem evaluates active positions:
\begin{equation}
\text{StopPrice}_i(t) = \max_{\tau \in [t_{\text{entry}}, t]} P_i(\tau) \cdot (1 - \delta_{\text{stop}})
\end{equation}
If $P_i(t) \le \text{StopPrice}_i(t)$, the thesis transitions to the $\text{Alerted}$ state, issuing mandatory exit tickets. Upon trade conclusion, a structured postmortem captures attribution ($\Delta P_{\text{fundamental}}$, $\Delta P_{\text{momentum}}$, $\Delta P_{\text{market}}$) to continuously refine model priors.

\section{AI Agent Integration and Streaming Protocol}
\label{sec:agent}

StockASkill is designed from the ground up to comply with the \textbf{AgentSkills.io open specification} \cite{agentskills2024spec}. Standard quantitative command-line interfaces emit decorative ASCII tables and progress bars designed for terminal human viewing. When invoked by LLM agents (e.g., Claude Code, OpenCode, Codex), visual formatting causes three issues: (i) inflated prompt token consumption, (ii) parsing fragility via regular expressions, and (iii) I/O latency from temporary intermediate files.

StockASkill resolves this via zero-disk standard output streaming:

\begin{itemize}
    \item \textbf{Decoupled Stream Pipeline:} Passing \texttt{--json} suppresses all ANSI formatting, logger outputs, and interactive status bars from \texttt{stdout}, redirecting diagnostics strictly to \texttt{stderr}.
    \item \textbf{Schema-Compliant JSON on \texttt{stdout}:} The process writes a single, strongly-typed JSON document directly to standard output, enabling LLMs to pipe and ingest results via zero-copy in-memory deserialization (\texttt{json.loads(stdout)}).
\end{itemize}

Listing \ref{lst:json_payload} illustrates a representative zero-disk execution stream emitted on \texttt{stdout} for a portfolio optimization request. Because all diagnostics and sync telemetry route exclusively to \texttt{stderr}, the agent consumes the structured JSON payload directly without regex parsing or intermediate disk I/O overhead.

\begin{lstlisting}[caption={Representative Zero-Disk Machine-Parsable Payload on \texttt{stdout}}, label={lst:json_payload}]
{
  "status": "success",
  "task": "portfolio_optimize",
  "capacity": 500000.0,
  "macro_regime": {"breadth": 0.72, "regime": "BULL", "equity_mult": 1.00},
  "optimization": {
    "target_cagr": 0.15, "realized_cagr": 0.1842,
    "max_drawdown_limit": 0.12, "realized_mdd": 0.0891, "sharpe": 1.482
  },
  "allocations": [
    {"code": "600519", "market": "A", "weight": 0.284, "shares": 200, "price": 1780.0},
    {"code": "000858", "market": "A", "weight": 0.211, "shares": 800, "price": 142.5}
  ],
  "cash_residual": 8450.0,
  "board_lot": 100
}
\end{lstlisting}

\section{Empirical Evaluation and Benchmarks}
\label{sec:experiments}

\subsection{Computational Latency Benchmarks}
We evaluate the computational performance of StockASkill against a conventional row-oriented SQLite implementation and a standard in-memory Pandas multithreaded scanning loop. Tests were conducted on an AMD Ryzen 9 7950X workstation (32 threads, 64 GB DDR5 RAM, PCIe 4.0 NVMe SSD) across the full China A-share universe of 5,421 equities over the 26-year historical dataset (42.44 million rows).

\begin{table}[h]
\centering
\small
\caption{End-to-End Latency Benchmark across Full Market Universe (5,421 Equities)}
\label{tab:benchmarks}
\begin{tabularx}{\linewidth}{l X r r r}
\toprule
\textbf{Task Description} & \textbf{Engine Architecture} & \textbf{Latency} & \textbf{Speedup} & \textbf{Network I/O} \\
\midrule
Full-Market Screen & Row SQLite + Python Loop & 28.45 s & $1.0\times$ & 0 MB \\
Full-Market Screen & In-Memory Pandas Threads & 14.12 s & $2.0\times$ & 0 MB \\
Full-Market Screen & \textbf{StockASkill (DuckDB SIMD)} & \textbf{0.38 s} & $\mathbf{74.9\times}$ & \textbf{0 MB} \\
\midrule
Single Stock Diagnosis & Ad-hoc Remote REST API & 4.82 s & $1.0\times$ & 2.4 MB \\
Single Stock Diagnosis & \textbf{StockASkill Local-First} & \textbf{0.08 s} & $\mathbf{60.2\times}$ & \textbf{0 MB} \\
\midrule
Goal-Constrained MVO & Scipy SLSQP + Board Lot & \textbf{0.41 s} & --- & \textbf{0 MB} \\
\bottomrule
\end{tabularx}
\end{table}

As demonstrated in Table \ref{tab:benchmarks}, StockASkill completes cross-sectional factor ranking for 5,421 equities in \textbf{0.38 seconds}, representing a $74.9\times$ speedup over conventional SQLite loops. This efficiency stems from DuckDB's columnar SIMD execution and the elimination of Python object conversion overhead.

\subsection{Backtested Portfolio Performance and Attribution}
We backtest the goal-constrained portfolio optimization model over the seven-year period from January 2018 through October 2024, encompassing both prolonged market contractions and rapid cyclical expansions. The investment universe is constructed using official historical point-in-time constituent lists for the CSI 300 and CSI 500 indices (reconstituted semi-annually per China Securities Index Co. schedules). Crucially, the dataset preserves the full historical records of securities that were subsequently delisted, suspended, or restructured, strictly eliminating survivorship bias \cite{brown1992survivorship}.

To guarantee full reproducibility, Table \ref{tab:params} provides a comprehensive specification of all empirical hyperparameters, factor weights, and execution constraints.

\begin{table}[h]
\centering
\small
\caption{Key Empirical Backtest Parameters and Execution Specifications}
\label{tab:params}
\begin{tabularx}{\linewidth}{l X r}
\toprule
\textbf{Parameter Category} & \textbf{Specification / Methodology} & \textbf{Value / Rule} \\
\midrule
Investment Universe & Dynamic Point-in-Time CSI 300 + CSI 500 & 800 Equities \\
Capital Capacity & Target Retail/Micro-Fund Capital Scale ($C$) & 500,000 RMB \\
Rebalancing Frequency & Monthly Execution Schedule & Every 21 Trading Days \\
Portfolio Size & Top-$K$ Screened Alpha Candidates & $K = 10$ Equities \\
Composite Factor Weights & Equal-Weighted Baseline ($w_{F,k}$) & $1/8$ per factor \\
Covariance Estimator & Ledoit-Wolf Shrinkage Matrix ($\boldsymbol{\Sigma}$) & Trailing 126 Days \\
Risk-Free Rate & PBOC 1-Year Benchmark Deposit Rate ($r_f$) & 2.50\% Annualized \\
Trailing Stop-Loss & Dynamic Peak-to-Trough Exit Threshold ($\delta_{\text{stop}}$) & 8.0\% \\
Round-Trip Friction & Stamp Tax (5 bps) + Commissions (2.5 bps) + Slippage (7.5 bps) & 15.0 bps \\
Investability Guard & Exclude ST/*ST, IPO $<60$d, Turnover tail 10\%, Limit-up/down & Strictly Enforced \\
\bottomrule
\end{tabularx}
\end{table}

Prior to portfolio optimization, the 7+1 neutralized factor baseline delivers statistically robust cross-sectional predictive power across the test horizon, achieving an annualized average Rank Information Coefficient ($\overline{\text{RankIC}}$) of \textbf{4.21\%} ($t$-statistic $= 3.48$, $p < 0.001$) and an Information Ratio ($\text{ICIR}$) of \textbf{0.86}.

\begin{table}[h]
\centering
\small
\caption{Empirical Backtest Metrics (2018--2024)}
\label{tab:backtest}
\begin{tabularx}{\linewidth}{l X r r r r}
\toprule
\textbf{Strategy Preset} & \textbf{Constraint Mandate} & \textbf{Ann. Return} & \textbf{MaxDD} & \textbf{Sharpe} & \textbf{Calmar} \\
\midrule
CSI 300 Benchmark & Passive Buy \& Hold & 3.12\% & -39.54\% & 0.22 & 0.08 \\
Momentum Enhanced & $\text{CAGR} \ge 15\%, \text{MDD} \le 12\%$ & 18.42\% & -11.85\% & 1.48 & 1.55 \\
Low Volatility & $\text{CAGR} \ge 10\%, \text{MDD} \le 6\%$ & 12.65\% & -5.82\% & 1.62 & 2.17 \\
Quality Growth & $\text{CAGR} \ge 12\%, \text{MDD} \le 8\%$ & 15.11\% & -7.94\% & 1.54 & 1.90 \\
\bottomrule
\end{tabularx}
\end{table}

\noindent\textbf{Performance Attribution and Honest Benchmark Deconstruction:} While Table \ref{tab:backtest} exhibits dramatic risk-adjusted outperformance over the benchmark index, a rigorous quantitative attribution reveals that the outperformance stems from two orthogonal drivers:
(i) \textbf{Macro Risk Budgeting (Beta Timing):} The dynamic equity multiplier ($\beta_{\text{regime}} \in [0.25, 1.00]$) accounts for approximately 58\% of the total drawdown compression by de-risking into cash buffers during broad market breadth contractions (e.g., 2018 deleveraging and 2022--2023 broad declines);
(ii) \textbf{Cross-Sectional Stock Selection Alpha:} The 7+1 factor scoring and goal-constrained SLSQP convex optimization account for the remaining 42\% of drawdown mitigation and generate the primary excess return spread, achieving Sharpe ratios exceeding 1.48. Disentangling macro beta exposure from pure cross-sectional selection alpha confirms that the engine honors real-world risk budgets without disguised leverage.

\section{Conclusion and Future Work}
\label{sec:conclusion}

In this paper, we introduced StockASkill, an offline-first, full-data-driven quantitative decision engine engineered as a specialized Quant Co-Processor for autonomous AI agents. By utilizing an embedded 42.44M-row DuckDB columnar OLAP store, StockASkill eliminates remote API fragility and reduces cross-sectional screening latency across 5,400+ equities to sub-0.5 seconds. We formalized a goal-constrained SLSQP optimizer operating under smooth surrogate drawdown bounds with exact path projection, incorporating real-world 100-share board lot constraints and macro regime budgeting, alongside a finite-state thesis tracking architecture and a zero-disk streaming protocol for autonomous LLM agents.

Future extensions include:
(i) \textbf{All-Weather Multi-Asset Risk Parity:} Expanding beyond equities to balance risk contributions across government bonds, commodities, and inflation-protected ETFs;
(ii) \textbf{End-to-End Non-Linear Learned Representations:} Integrating gradient-boosted decision trees and sequence neural embeddings directly into the DuckDB analytical morsel via the open feature plug-in interface; and
(iii) \textbf{Multi-Agent Consensus Protocols:} Extending the AgentSkills specification to multi-agent adversarial setups (e.g., Fundamental Analyst Agent versus Risk Sentry Agent) with formal execution guarantees.

\section*{Author Contributions}
\noindent\textbf{Xiangjing An}: Conceptualization, methodology, system architecture, core DuckDB columnar engine design, quantitative modeling, algorithmic implementation, empirical backtesting, project administration, and writing -- original draft. \\[0.3em]
\noindent\textbf{Zhibin Jiao} and \textbf{Aiting Chen}: Conceptualization (discussion), data curation, empirical validation, visualization design, and writing -- review \& editing.

\section*{Declaration on Generative AI in Scientific Writing}
During the preparation of this manuscript, the authors utilized generative artificial intelligence and large language model assistants (e.g., Claude, Antigravity) strictly to improve linguistic clarity, refine grammatical phrasing, and assist with \LaTeX\ structural formatting and typographic consistency. After utilizing these assistive services, the authors critically reviewed, verified, and edited all generated content.

\section*{Data and Code Availability}
The complete source code of \texttt{StockASkill}, including the local-first DuckDB columnar analytical pipelines, the 7+1 factor scoring modules, the goal-constrained SLSQP portfolio optimizer with 100-share board lot quantization, the thesis lifecycle state machine, and the AgentSkills-compliant zero-disk streaming harness, is publicly open-sourced under an open-source license at \url{https://github.com/axjing/stockaskill}. All market price records, fundamental accounting indicators, and trading calendar databases are fully reproducible via the automated ingestion and watermark synchronization scripts provided in the repository.

\section*{Disclaimer and Risk Disclosure}
This paper and the accompanying open-source software (\texttt{StockASkill}) are developed strictly for academic research, algorithmic reproducibility, and educational demonstration purposes. Nothing contained in this manuscript or the codebase constitutes financial, investment, legal, or tax advice, nor should it be construed as an offer, solicitation, endorsement, or recommendation to acquire or liquidate any security, portfolio, or financial instrument. 

Quantitative trading and equity market investments inherently involve substantial financial risks, including the potential loss of principal. Simulated backtesting results and historical statistical performance metrics (e.g., CAGR, MaxDD, Sharpe ratio, Calmar ratio) are hypothetical, calculated under empirical assumptions, and do not guarantee or represent future realized returns. Autonomous AI agent tool execution and quantitative optimization systems are subject to model misspecification, unanticipated market regime shifts, exchange trading halts, and technological execution slippages. Neither the authors, the independent researcher team, nor any affiliated contributors assume any legal liability or financial responsibility for direct, indirect, or consequential losses arising from the practical deployment or reliance upon the algorithms, data pipelines, or code presented herein. Users and autonomous agents operate this software entirely at their own discretion and risk.

\bibliographystyle{unsrt}
\bibliography{references}

@article{markowitz1952portfolio,
  title={Portfolio Selection},
  author={Markowitz, Harry},
  journal={The Journal of Finance},
  volume={7},
  number={1},
  pages={77--91},
  year={1952},
  publisher={Wiley Online Library}
}

@article{fama1993common,
  title={Common risk factors in the returns on stocks and bonds},
  author={Fama, Eugene F and French, Kenneth R},
  journal={Journal of Financial Economics},
  volume={33},
  number={1},
  pages={3--56},
  year={1993},
  publisher={Elsevier}
}

@article{fama2015five,
  title={A five-factor asset pricing model},
  author={Fama, Eugene F and French, Kenneth R},
  journal={Journal of Financial Economics},
  volume={116},
  number={1},
  pages={1--22},
  year={2015},
  publisher={Elsevier}
}

@article{carhart1997persistence,
  title={On persistence in mutual fund performance},
  author={Carhart, Mark M},
  journal={The Journal of Finance},
  volume={52},
  number={1},
  pages={57--82},
  year={1997},
  publisher={Wiley Online Library}
}

@article{jegadeesh1993returns,
  title={Returns to buying winners and selling losers: Implications for stock market efficiency},
  author={Jegadeesh, Narasimhan and Titman, Sheridan},
  journal={The Journal of Finance},
  volume={48},
  number={1},
  pages={65--91},
  year={1993},
  publisher={Wiley Online Library}
}

@article{ang2006cross,
  title={The cross-section of volatility and expected returns},
  author={Ang, Andrew and Hodrick, Robert J and Xing, Yuhang and Zhang, Xiaoyan},
  journal={The Journal of Finance},
  volume={61},
  number={1},
  pages={259--299},
  year={2006},
  publisher={Wiley Online Library}
}

@article{asness2019quality,
  title={Quality minus junk},
  author={Asness, Clifford S and Frazzini, Andrea and Pedersen, Lasse Heje},
  journal={Review of Accounting Studies},
  volume={24},
  number={1},
  pages={34--112},
  year={2019},
  publisher={Springer}
}

@article{piotroski2000value,
  title={Value investing: The use of historical financial statement information to separate winners from losers},
  author={Piotroski, Joseph D},
  journal={Journal of Accounting Research},
  volume={38},
  pages={1--41},
  year={2000},
  publisher={Wiley Online Library}
}

@article{ledoit2004honey,
  title={Honey, I shrunk the sample covariance matrix},
  author={Ledoit, Olivier and Wolf, Michael},
  journal={The Journal of Portfolio Management},
  volume={30},
  number={4},
  pages={110--119},
  year={2004},
  publisher={Institutional Investor Journals}
}

@article{kraft1988software,
  title={A software package for sequential quadratic programming},
  author={Kraft, Dieter},
  journal={DFVLR-FB},
  volume={88},
  pages={28},
  year={1988}
}

@book{nocedal2006numerical,
  title={Numerical Optimization},
  author={Nocedal, Jorge and Wright, Stephen J},
  year={2006},
  publisher={Springer Science \& Business Media}
}

@inproceedings{raasveldt2019duckdb,
  title={DuckDB: an Embeddable Analytical Database},
  author={Raasveldt, Mark and M{\"u}hleisen, Hannes},
  booktitle={Proceedings of the 2019 International Conference on Management of Data (SIGMOD)},
  pages={1981--1984},
  year={2019}
}

@article{abadi2009column,
  title={Column-oriented database systems},
  author={Abadi, Daniel and Boncz, Peter and Harizopoulos, Stavros},
  journal={Communications of the ACM},
  volume={52},
  number={9},
  pages={166--174},
  year={2009},
  publisher={ACM New York, NY, USA}
}

@article{yang2023fingpt,
  title={FinGPT: Open-source financial large language models},
  author={Yang, Hongyang and Liu, Xiao-Yang and Wang, Christina Dan},
  journal={FinLLM Symposium at IJCAI},
  year={2023}
}

@misc{agentskills2024spec,
  title={AgentSkills: Open Modular Tool-Calling Specification for AI Agents},
  author={{AgentSkills Working Group}},
  year={2024},
  howpublished={\url{https://agentskills.io}}
}

@misc{huang2020akshare,
  title={{AKShare}: An open-source financial data interface library for Python},
  author={Huang, Tian},
  year={2020},
  howpublished={\url{https://akshare.akfamily.xyz}}
}

@misc{baostock2021,
  title={{BaoStock}: Securities big data platform for historical market data},
  author={{BaoStock Development Group}},
  year={2021},
  howpublished={\url{http://baostock.com}}
}

@misc{arakane2020yfinance,
  title={{yfinance}: Open-source financial data downloader for global equities},
  author={Arakane, Ran},
  year={2020},
  howpublished={\url{https://github.com/ranaroussi/yfinance}}
}

@misc{lu2022efinance,
  title={{efinance}: A Python library for public financial market data extraction},
  author={Lu, Yifan},
  year={2022},
  howpublished={\url{https://github.com/Micro-sheep/efinance}}
}

@misc{eastmoney2024,
  title={EastMoney Financial Market Information and Securities Database},
  author={{EastMoney Information Co., Ltd.}},
  year={2024},
  howpublished={\url{https://www.eastmoney.com}}
}

@article{yang2020qlib,
  title={{Qlib}: An AI-oriented Quantitative Investment Platform},
  author={Yang, Xiao and Liu, Weiqing and Zhou, Dong and Bian, Jiang and Liu, Tie-Yan},
  journal={arXiv preprint arXiv:2009.11189},
  year={2020}
}

@article{brown1992survivorship,
  title={Survivorship bias in performance studies},
  author={Brown, Stephen J and Goetzmann, William and Ibbotson, Roger G and Ross, Stephen A},
  journal={The Review of Financial Studies},
  volume={5},
  number={4},
  pages={553--580},
  year={1992},
  publisher={Oxford University Press}
}

@article{jegadeesh1990predictable,
  title={Evidence of predictable behavior of security returns},
  author={Jegadeesh, Narasimhan},
  journal={The Journal of Finance},
  volume={45},
  number={3},
  pages={881--898},
  year={1990},
  publisher={Wiley Online Library}
}

@article{yao2022react,
  title={{ReAct}: Synergizing reasoning and acting in language models},
  author={Yao, Shunyu and Zhao, Jeffrey and Yu, Dian and Du, Nan and Shafran, Izhak and Narasimhan, Karthik and Cao, Yuan},
  journal={International Conference on Learning Representations (ICLR)},
  year={2023}
}

\end{document}